\documentclass[letterpaper, 10 pt, conference]{ieeeconf}

\usepackage{graphicx}
\usepackage{subfig}
\usepackage{amsmath}
\usepackage{amssymb}
\usepackage{booktabs}
\usepackage{tabularx}
\usepackage{multicol}
\usepackage{multirow}
\usepackage{bm}
\usepackage{xcolor} 
\usepackage{cite}

\usepackage{tablefootnote}

\usepackage{caption}
\usepackage[ruled]{algorithm2e}

\newcommand{\point}{\bm{p}}

\newcommand{\normal}{\bm{n}}

\newcommand{\plane}{\mathcal{P}}

\newcommand{\node}{\mathcal{N}}

\newcommand{\points}{P}

\newcommand{\datamatrix}{\bm{X}}
\newcommand{\mean}{\bm{\mu}}
\newcommand{\onematrix}{\bm{1}}

\newcommand{\trans}{\mathrm{T}}

\newcommand{\PlaneList}{L_\mathcal{P}}
\newcommand{\NodesQueue}{Q_\mathcal{N}}
\newcommand{\PointsList}{L_p}

\newcommand{\realR}{\mathbb{R}}
\newcommand{\inlierNumber}{N_\text{in}}

\newcommand{\gravity}{\bm{g}}

\IEEEoverridecommandlockouts 

\title{\LARGE \bf Multi-Resolution Planar Region Extraction for Uneven Terrains}

\author{Yinghan Sun$^{1}$, Linfang Zheng$^{1, 2}$, Hua Chen$^{1}$ and Wei Zhang$^{1}$
\thanks{$^{1}$Shenzhen Key Laboratory of Control Theory and Intelligent Systems, School of System Design and Intelligent Manufacturing, Southern University of Science and Technology, Shenzhen, China.
Emails:
        {\tt\footnotesize \{sunyh2021, 11956001\}@mail.sustech.edu.cn, \{chenh6, wzhang3\}@sustech.edu.cn}}%
\thanks{$^{2}$School of Computer Science, University of Birmingham, UK.
        {\tt\footnotesize lxz948@student.bham.ac.uk}}%
}

\begin{document}

\maketitle
\thispagestyle{empty}
\pagestyle{empty}

\begin{abstract}
This paper studies the problem of extracting planar regions in uneven terrains from unordered point cloud measurements. Such a problem is critical in various robotic applications such as robotic perceptive locomotion. While existing approaches have shown promising results in effectively extracting planar regions from the environment, they often suffer from issues such as low computational efficiency or loss of resolution. To address these issues, we propose a multi-resolution planar region extraction strategy in this paper that balances the accuracy in boundaries and computational efficiency. Our method begins with a pointwise classification preprocessing module, which categorizes all sampled points according to their local geometric properties to facilitate multi-resolution segmentation. Subsequently, we arrange the categorized points using an octree, followed by an in-depth analysis of nodes to finish multi-resolution plane segmentation. The efficiency and robustness of the proposed approach are verified via synthetic and real-world experiments, demonstrating our method's ability to generalize effectively across various uneven terrains while maintaining real-time performance, achieving frame rates exceeding 35 FPS.
\end{abstract}
\section{Introduction}
Legged robots are capable of traversing complex and challenging terrains. This unique ability makes them well-suited for missions such as rescue tasks, where wheeled robots of comparable size face substantial difficulties. In recent years, significant progress has been achieved in legged locomotion~\cite{rl_perceptive_locomotion_sr2022, anymal_parkour_preprint2023, perceptive_locomotion_wbc_ral2020, perceptive_locomotion_traj_opti_2021icra, perceptive_mixed_integer_humanoids2014, perceptive_mixed_integer_iros2017, perceptive_mixed_integer_ral2017, perceptive_a_star_humanoids2019, perceptive_wbc_ransac_iros2019, wenchun_perceptive_stair_iros2021, perceptive_mpc_plane_icra2021, perceptive_nonlinear_mpc_tro2023}. Despite these advancements, it remains challenging to control robots to traverse complex and uneven terrains effectively. One promising pipeline in recent methods~\cite{perceptive_mixed_integer_humanoids2014, perceptive_mixed_integer_iros2017, perceptive_mixed_integer_ral2017, perceptive_a_star_humanoids2019, perceptive_wbc_ransac_iros2019, wenchun_perceptive_stair_iros2021, perceptive_mpc_plane_icra2021, perceptive_nonlinear_mpc_tro2023} is incorporating geometric visual information into legged locomotion, which segments the input visual data into multiple planes and then uses this information to plan the robot's footholds. However, achieving precise and efficient plane segmentation, particularly in uneven terrains, continues to pose challenges.

\begin{figure}
\centering
\vspace{2.5mm}
\includegraphics[width=\linewidth, page=1, trim = 80 70 72 68, clip]{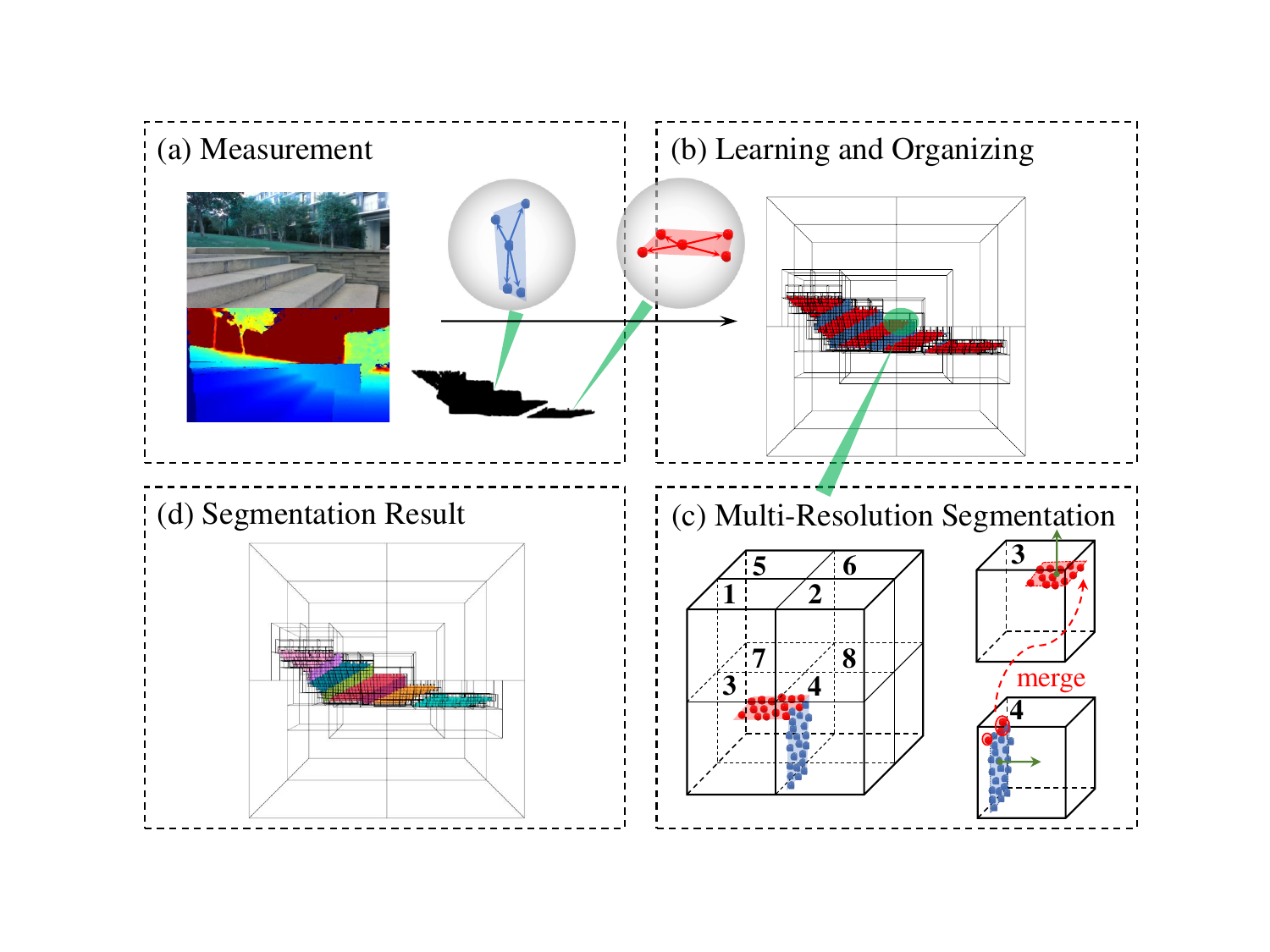}
\caption{\textbf{The illustration of the proposed method's key idea.} (a) The measured point cloud. (b) Pointwise classification by learning their local geometric properties and organization using an octree. Each square represents an octree's node. (c) Multi-resolution plane segmentation. Children 1 to 2 and 5 to 8 have no points and are therefore not visited. (d) Segmentation results, with points in color belonging to the same plane.}
\label{fig:teaser}
\vspace{-7mm}
\end{figure}
In this paper, we introduce a novel approach for efficient, accurate, and robust plane segmentation applicable to uneven terrains. Our method initiates by preprocessing the input point clouds via a pointwise classification module, which categorizes points into a number of groups based on local surface inclinations. We then organize the unordered point clouds via an octree~\cite{octree_1980}, incorporating a specially designed multi-resolution structure.
To achieve multi-resolution segmentation, we employ the divide-and-conquer strategy, capitalizing on an in-depth analysis of point distribution within each node. This process is facilitated by a set of proposed criteria that ensure efficiency and noise robustness. In particular, to avoid redundant and time-consuming computations when merging coplanar planes, we propose an incremental implementation for updating the point cluster's covariance matrix. Experimental results demonstrate the robustness of our method across diverse terrains and noise levels, all while achieving real-time performance.

\subsection{Related Work}
The research on plane segmentation has attracted considerable attention and advanced progressively in recent years. They can be roughly divided into the Random Sample Consensus (RANSAC)~\cite{ransac_1981} based methods~\cite{ransac_detection_2007, perceptive_a_star_humanoids2019, ransac_bipedal_iros2021, perceptive_mpc_plane_icra2021, octree_RANSAC_iros2020} and the region growing-based methods~\cite{pointwise_region_growing_iros2008, pointwise_region_growing_ias2012, peac_icra2014, cape_plane_extraction_2018iros}.

\textit{\textbf{RANSAC-based Methods.}} 
RANSAC-based methods usually use RANSAC iteratively on unordered point clouds and remove inliers corresponding to the extracted plane instances. Such a procedure has demonstrated an exceptional ability in handling outliers and successfully segment planes in various legged visual locomotion works~\cite{perceptive_a_star_humanoids2019, ransac_bipedal_iros2021, perceptive_mpc_plane_icra2021}. 
A notable variation is presented in \cite{octree_RANSAC_iros2020}, where an octree is employed to structure the unordered point cloud, reducing the need for multiple RANSAC iterations. However, this approach demonstrates speed limitations due to the expensive computation.
Moreover, RANSAC exhibits a greedy nature, meaning that any incorrect or missed matches in earlier models cannot be rectified later~\cite{cv_book_2012}.

\textit{\textbf{Region-growing-based Methods.}} These methods operate by initially identifying local planar regions and then iteratively expanding them through merging coplanar regions. Early works~\cite{pointwise_region_growing_iros2008, pointwise_region_growing_ias2012} primarily focused on pointwise region-growing, demonstrating strong segmentation performance. However, they tend to be time-consuming due to the computational overhead of pointwise searching and merging. Recent advancements in region-growing-based techniques~\cite{peac_icra2014, cape_plane_extraction_2018iros, zhixu_plane_extraction_2022iros} have enhanced processing speed by incorporating connectivity information extracted from 2D image patches and merging coplanar regions to achieve real-time performance. Nevertheless, operating on image patches may sacrifice the plane segmentation resolution, potentially resulting in imprecise plane boundaries~\cite{peac_icra2014, cape_plane_extraction_2018iros, zhixu_plane_extraction_2022iros}.

\subsection{Contributions}
The main contributions of this paper are summarized as follows:
1) We propose a multi-resolution plane segmentation method that achieves a balance between accuracy and efficiency to meet the requirement of practical applicability. 
Our method significantly outperforms the RANSAC-based approach, achieving nearly 10 times the processing speed. Moreover, it exhibits a notable reduction of missing points across diverse noise levels compared to region-growing-based methods.
2) We propose a local geometric sensitive pointwise classification module, which reduces the average error on the normal vectors of the extracted planes by approximately tenfold and demonstrates remarkable noise robustness. It also exhibits impressive performance on real-world uneven terrains.
3) We introduce an updating scheme for the covariance matrix estimation that incrementally incorporates new data points during coplanar regions merging. This strategy eliminates redundant matrix multiplication, enhancing the overall effectiveness of the proposed method.
\section{Problem Description and Overview}
\label{sec:problem_description}

\subsection{Problem Description and Notation Conventions}
This paper studies the plane segmentation problem of identifying planes from perceptive measurements. In particular, we consider the input as the 3D point cloud. For depth camera-based applications, the point cloud measurement is obtained by back-projecting the depth image using camera parameters. Specifically, the objective of the plane segmentation problem in this paper is to partition the input points into distinct subsets, each adhering to specific planarity criteria outlined below. The final output is a list of planes, denoted as $\PlaneList$, composed of these subsets. 

In this paper, we represent a plane $\plane$ using a tuple $(P, N, \mean, \normal, e_n)$, where $\points$ is a set containing all the points in the plane, $N$ and $\mean \in \realR^3$ are the point number and centroid of $\points$, $\normal \in \realR^3$ is the plane's normal vector, and $e_n \in \realR$ is the mean square error (MSE) in the normal direction. 

\subsection{Overview of the Proposed Framework}
\begin{figure*}[htbp]
\centering
\includegraphics[width=\textwidth, trim = 55 110 47 135, clip]{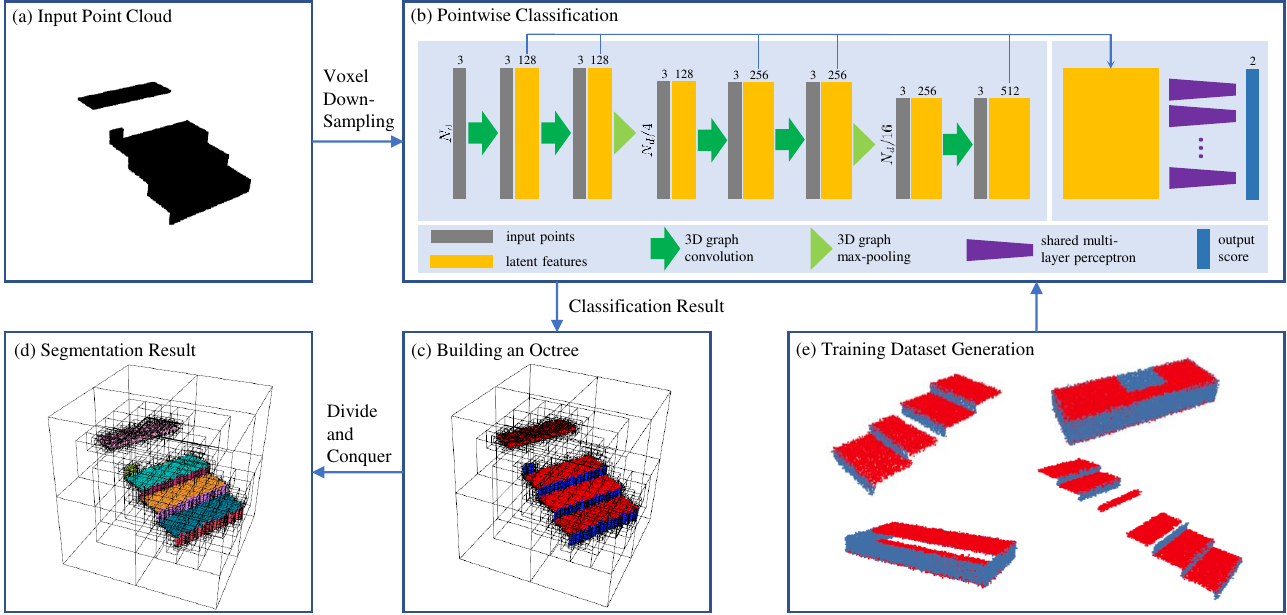}
\caption{\textbf{The overall framework of the proposed method.} (a) We perform voxel down-sampling on the input point cloud to obtain a point cloud with size $N_d$ for post-processing. (b) The down-sampled points are categorized based on the local surface's inclination around each point. The diagram shows the structure of the 3D-GCN utilized in this step. (c) We organize all points along with their labels using an octree. Each cube represents a node within the octree, with larger cubes representing nodes closer to the root. (d) We apply the divide-and-conquer strategy to achieve multi-resolution plane segmentation. Points with the same color belong to the same plane. (e) To train the neural network, we create a small and simple dataset with proper data augmentations.}
\label{fig:framework}
\vspace{-6mm}
\end{figure*}

The schematic diagram of our method is given in Figure~\ref{fig:framework}. Initially, the input point cloud undergoes voxel downsampling to yield a fixed size of $N_d$ points. Subsequently, a pointwise classification step separates points into two categories based on their local surface inclination. These classified points are then organized using an octree, where we assess different point distribution cases within the octree nodes by utilizing point categorical labels. Finally, we implement the multi-resolution plane segmentation using the divide-and-conquer strategy by traversing the octree.

\section{Pointwise Classification}
To assist our multi-resolution plane segmentation framework, we incorporate a pointwise classification module to pre-process the data. We apply the 3D graph convolution network (3D-GCN)~\cite{3dgcn_cvpr2020} to categorize each point based on its local surface inclinations, thanks to its ability to extract local geometric features from unordered point clouds with shift and scale-invariant properties.

Broadly speaking, points can be classified into distinct categories based on the angle $\beta_i$ between their local surface normal $\normal_{l,i}$ and selected baseline vectors $\bm{b} \in \realR^3$:
\begin{equation}
    \beta_i = \arccos \left( \frac{|\bm{b}^\trans \bm{n}_{l,i}|}{\|\bm{b}\|\|\bm{n}_{l,i}\|} \right).
\end{equation}
In this work, we exclusively use the gravity vector $\gravity$ as the baseline vector, resulting in a classification into two categories. Specifically, a point is categorized as $h$ if $\beta_i \leq \frac{\pi}{4}$; otherwise, it is assigned to category $v$.

To train the classification module, we propose to generate a synthetic dataset (as depicted in Figure~\ref{fig:framework}(e)), which contains three types of shapes. Each shape is defined by parameters sampled uniformly from specified ranges: 200 staircases with step dimensions varying from [0.2, 0.4]m in length, [0.2, 1.5]m in width, and [0.08, 0.3]m in height; 200 boxes with dimensions ranging from [0.1, 2.0]m in length and width and [0.08, 0.3]m in height; and 200 planes with dimensions ranging from [0.1, 2.0]m in length and width, and unit normal vectors in the $z$-component varying from [-1, 1]. To enhance the noise robustness of the trained classifier, we apply the following augmentations to the simulated data: 1) Uniform disturbance of point coordinates ranging from [-0.01, 0.01]m. 2) Random rotation along the $z$-axis. 3) Removing random clusters from the original point cloud. The network is therefore trained via such a simple dataset to classify each point into two categories.

\section{Plane Segmentation}

In this section, we first introduce the design of the nodes' structure in an octree. We then delve into the analysis of different point distribution cases within a node. This analysis will serve as a guide for determining when to divide, when to conquer, and how to conquer a node. Subsequently, we outline the multi-resolution plane segmentation process through octree traversal. Additionally, an incremental implementation for plane merging is proposed to accelerate the process.

\subsection{The Structure of the Octree}
An octree~\cite{octree_1980} is a hierarchical tree structure used to organize 3D data. Each node in the octree has eight children. The octree's spatial resolution is determined by its maximum depth and the input point cloud's spatial size. To assist the plane segmentation task, each node, denoted as $\node$, stores essential spatial information. This includes its depth in the octree, the spatial bounding box defined by two corner points, pointers to its children and ancestor, counts of points labeled as $h$ and $v$ in the node, and a set $\points$ containing all points within the node. Notably, the octree can be efficiently constructed using binary insertion, which effectively manages unordered points.

\subsection{Node Analysis}
\begin{figure}[htbp]
\vspace{-3mm}
\centering
\includegraphics[width=\linewidth, page=1, trim = 23 200 130 135, clip]{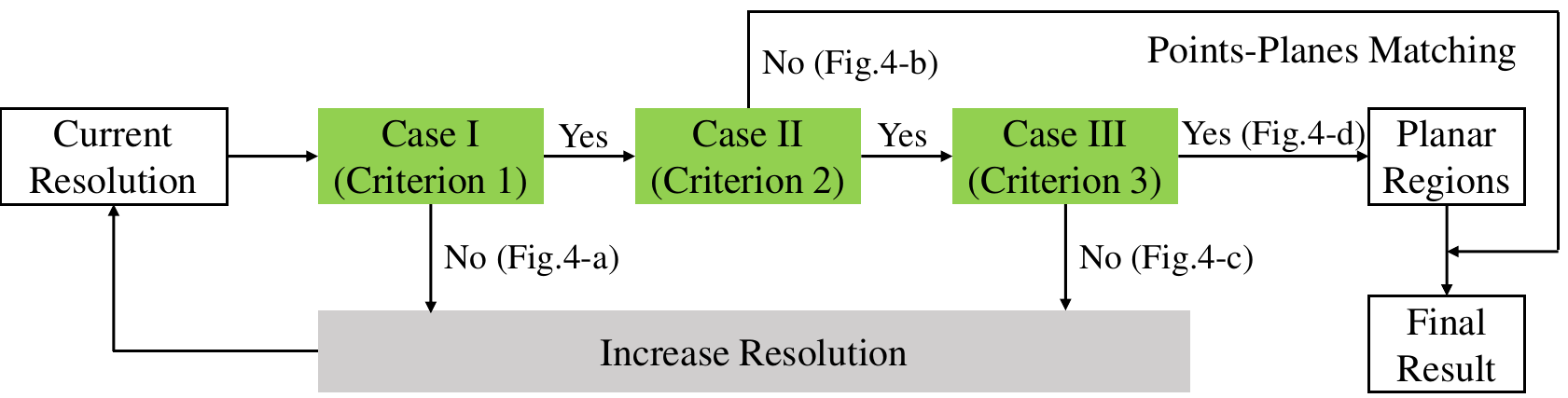}
\caption{\textbf{Schematic Diagram of the Multi-Resolution Plane Segmentation Process.} The green blocks represent the decision box.}
\label{fig:procedure}
\vspace{-2mm}
\end{figure}
In this subsection, we analyze the different cases that can arise in a node. By applying specific criteria, we aim to determine the optimal moments for node division and conquering and the strategies employed during conquering. This approach enables us to segment planes at multiple resolutions effectively. For clarity, we define the inlier points of a node as the majority of points sharing the same label, while the remaining points are classified as outliers.
\begin{figure}[htbp]
\centering
	\begin{minipage}{0.24\linewidth}
		\centering
		\includegraphics[width=0.98\linewidth, page=3, trim = 280 200 280 180, clip]{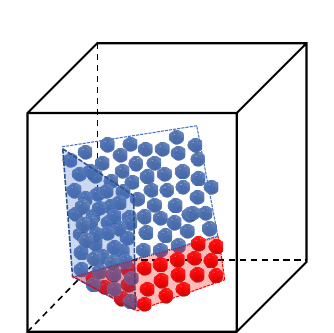}
	\end{minipage}
	\begin{minipage}{0.24\linewidth}
		\centering
		\includegraphics[width=0.98\linewidth, page=2, trim = 280 200 280 180, clip]{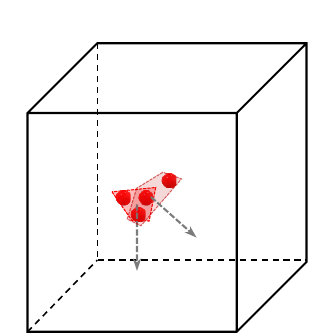}
        \end{minipage}
	\begin{minipage}{0.24\linewidth}
		\centering
		\includegraphics[width=0.98\linewidth, page=4, trim = 280 200 280 180, clip]{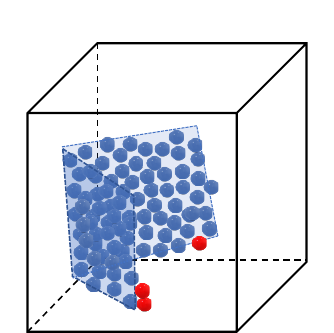}
	\end{minipage}
	\begin{minipage}{0.24\linewidth}
		\centering
		\includegraphics[width=0.98\linewidth, page=5, trim = 280 200 280 180, clip]{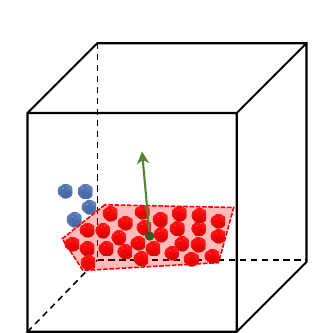}
        \end{minipage}

        \vspace{2mm}
        \begin{minipage}{0.24\linewidth}
            \centering
            \footnotesize{(a)}
        \end{minipage}
        \begin{minipage}{0.24\linewidth}
            \centering
            \footnotesize{(b)}
        \end{minipage}
        \begin{minipage}{0.24\linewidth}
            \centering
            \footnotesize{(c)}
        \end{minipage}
        \begin{minipage}{0.24\linewidth}
            \centering
            \footnotesize{(d)}
        \end{minipage}

\caption{\textbf{The illustration of different cases in a node.} {\color{red}Red} color represents the points labeled as $h$, and {\color{blue}blue} represents points labeled as $v$. (a) If a node contains an excessive number of outliers, it is likely to encompass multiple planes. In such cases, we expand the node without applying PCA to prevent unnecessary matrix multiplications. (b) Using too few points to fit a plane can lead to sensitivity to noise. (c) A node may contain multiple planes, even if it meets Criteria 1 and 2. In such cases, the resulting MSE ($e_n$) of the PCA tends to be large. (d) When all criteria are met, we consider the region constructed by the inlier points as a plane candidate. The green arrow and point indicate the plane's normal vector and centroid, respectively.
}
\label{fig:node_diff_cases}
\vspace{-5mm}
\end{figure}

\subsubsection{Case I} We evaluate a node's purity by examining the number of outliers ($N_\text{out}$) and the outlier ratio ($r_\text{out}$) using Criterion 1. If the criterion is not satisfied, as shown in Figure \ref{fig:node_diff_cases}(a), it is highly unlikely for a node to contain only one plane candidate. In such cases, we opt to divide the node.

\noindent \textbf{Criterion 1} (Purity Criterion) We say a node satisfies the purity condition if the following conditions hold:
\begin{equation}
        N_\text{out} \leq \theta_{N_\text{out}}, \text{ and } r_\text{out} \leq \theta_{r_\text{out}},
        \label{eq:purity_condition}
\end{equation}
where $\theta_{N_\text{out}}$ and $\theta_{r_\text{out}}$ are two corresponding thresholds.

\subsubsection{Case II}
To determine a plane's equation in space, a minimum of three points is required. However, relying on too few points can lead to undue influence from edges and noise, as demonstrated in Figure \ref{fig:node_diff_cases}(b). To enhance robustness against noise, 
we introduce a hyperparameter that governs the minimum number of points necessary to fit a plane, which leads to the following criterion:


\noindent \textbf{Criterion 2} (Minimum Inlier Points Criterion)
We say a node holds for the minimum inlier points criterion if the number of inlier points $N_\text{in}$ is larger than a threshold $\theta_{N_\text{in}}$:
\begin{equation}
    N_\text{in} \geq \theta_{N_\text{in}}.
    \label{eq:minimum_inlier_points_criteria}
\end{equation}

\subsubsection{Case III}
For nodes that satisfy Criterion 1 and Criterion 2, we conquer them following~\cite{cape_plane_extraction_2018iros} and extract their corresponding planes using the principal component analysis (PCA). More precisely, given the inliers' mean position $\mean_\text{in} \in \realR^3$ and data matrix $\datamatrix_\text{in} = [\point_1, \cdots, \point_{\inlierNumber}] \in \mathbb{R}^{3 \times \inlierNumber}$, we apply eigenvalue decomposition on the positive semi-definite covariance matrix $\bm{\Sigma}_\text{in} \in \realR^{3 \times 3}$:
\begin{equation}
    \bm{\Sigma}_\text{in} = (\datamatrix_\text{in} - \mean_\text{in} \onematrix)(\datamatrix_\text{in} - \mean_\text{in} \onematrix)^\trans,
\end{equation}
where $\onematrix \in \mathbb{R}^{1 \times \inlierNumber}$ is a one matrix. The smallest eigenvalue of $\bm{\Sigma}_\text{in}$ corresponds to the MSE in the normal direction $e_n$, and its corresponding eigenvector represents the normal vector $\normal$.
We use Criterion 3 to check whether these points belong to a single plane. If the criterion is not met, as illustrated in Figure \ref{fig:node_diff_cases}(c), we further divide the node to extract planes at a finer resolution. Otherwise, the plane is directly extracted from this node (see Figure~\ref{fig:node_diff_cases}(d)). Note that all the operations are conducted on the inlier points. We will show how to deal with the outliers in the next subsection.

\noindent \textbf{Criterion 3} (Plane Candidate Criterion) We accept a set containing all the inlier points in a node as a plane candidate if the MSE in the normal direction $e_n$ of these points satisfies
\begin{equation}
    e_n \geq \theta_{e_n}.
    \label{eq:plane_candidate_criteria}
\end{equation}

\subsection{Traversing the Octree}
After analyzing different cases in a node, we proceed to segment planes by traversing the octree. The key idea is to explore nodes that require division. We establish a queue $\NodesQueue$ to store nodes awaiting exploration. Upon visiting a dequeued node, we evaluate it based on the previously discussed cases. If the node is suitable for conquering, we derive plane parameters from its points. Otherwise, we either enqueue all its children into $\NodesQueue$ or add its points to a list $\PointsList$ if it's a leaf node, including any identified outliers from Case III. We go through each node until $\NodesQueue$ is empty. 

For each extracted plane candidate, we apply Criterion 4 to check whether a coplanar plane already exists in $\PlaneList$. If a coplanar plane is found, we merge the candidate plane with the existing one and update its parameters in $\PlaneList$ accordingly. An incremental implementation of this procedure is discussed in Section~\ref{sec:plane_merging}.

\noindent \textbf{Criterion 4} (Coplanarity Criterion) We say $\plane_1$ and $\plane_2$ are coplanar if their normal $\normal$ and mean position $\mean$ satisfy:
    \begin{equation}
        \left\{  
             \begin{array}{lr}  
             |\normal_1^\trans \normal_2| \geq \theta_{\text{coplane}} &  \\  
             \left|\normal_1^\trans (\mean_1 - \mean_2) / {\|\mean_1 - \mean_2\|}\right| \leq (1 - \theta_{\text{coplane}}) & \\  
             \left|\normal_2^\trans (\mean_1 - \mean_2) / {\|\mean_1 - \mean_2\|}\right| \leq (1 - \theta_{\text{coplane}}) &    
             \end{array}  
        \right.
    \end{equation}

Finally, we address the points in $\PointsList$. Note that a plane $\plane$ can be defined by its $\mean$ and $\normal$:
\begin{equation}
    n_x (x - \mu_x) + n_y (y - \mu_y) + z(z - \mu_z) = 0.
    \label{eq:plane_equation}
\end{equation}
Then, it is convenient to justify whether a point belongs to a plane by computing the distance from points to planes. We use a threshold on this distance that equals $0.005$m to handle the noise in the real case. After checking all the points in $\PointsList$, we get the final segmentation on the point cloud. 

\subsection{Plane Merging}\label{sec:plane_merging}
If a newly extracted plane $\plane_2$ is coplanar with an existing plane $\plane_1$, we merge them and update the plane's properties by applying PCA on the union of $\points_1$ and $\points_2$. However, computing the covariance matrix for this new point set can be computationally demanding, especially if either plane contains a large number of points. 

To accelerate this process, we extend the incremental implementation for covariance matrix updates from \cite{pointwise_region_growing_iros2008}, originally designed for pointwise updates, to handle point clusters. Specifically, we extend the plane representation in Section~\ref{sec:problem_description} by including an additional term $\datamatrix \datamatrix^\trans \in \mathbb{R}^{3 \times 3}$. This term contributes to the covariance matrix and can be updated incrementally, as demonstrated below:
\begin{equation}
    \datamatrix_1 \datamatrix_1^\trans \leftarrow \datamatrix_1 \datamatrix_1^\trans + \datamatrix_2 \datamatrix_2^\trans.
    \label{data_update}
\end{equation}

Secondly, $\mean$, $N$, and $\points$ could be updated easily by
\begin{equation}
    \mean_1 \leftarrow \frac{N_1}{N_1 + N_2} \mean_1 + \frac{N_2}{N_1 + N_2} \mean_2.
    \label{mean_update}
\end{equation}
\begin{equation}
    N_1 \leftarrow N_1 + N_2, \quad \points_1 \leftarrow \points_1 \cup \points_2.
\end{equation}

Note that the covariance matrix could be expressed as a linear combination of $\datamatrix \datamatrix^\trans$ and $\mean \mean^\trans$:
\begin{equation}
    \begin{aligned}
    & (\datamatrix - \mean \onematrix)(\datamatrix - \mean \onematrix)^\trans \\
    = & \datamatrix \datamatrix^\trans - \datamatrix \onematrix^\trans \mean^\trans - \mean \onematrix \datamatrix^\trans + \mean \onematrix \onematrix^\trans \mean^\trans \\
    = & \datamatrix \datamatrix^\trans - N \mean \mean^\trans.
    \end{aligned}
    \label{update_covariance}
\end{equation}

As $\datamatrix \datamatrix^\trans$, $N$, and $\mean$ in the plane representation are updated incrementally, the new covariance matrix can be efficiently computed using Equation (\ref{update_covariance}). Subsequently, $\normal$ and $e_n$ are updated through the eigenvalue decomposition of the updated $3 \times 3$ covariance matrix.
\section{Experiments}
We perform a series of ablation studies to evaluate the proposed method's precision, efficiency, and robustness quantitatively. We also test our method's generalizability on real-world collected data. 

\textbf{Implementation Details:}
Our octree implementation and related algorithms were developed in C++. All experiments were conducted on a computer equipped with an Intel(R) Core(TM) i9-10900K CPU, 32GB RAM, and a single NVIDIA GeForce RTX 3090 GPU. We adopt the official code from 3D-GCN~\cite{3dgcn_cvpr2020} with its recommended parameters for the pointwise classification module. The generated synthetic dataset used for evaluation comprises 100 stairs with varying sizes, heights, and numbers of steps. Each stair featured 2 to 6 steps, with step dimensions ranging from 0.2m to 0.4m in length, 0.3m to 1.5m in width, and 0.1m to 0.3m in height. We utilized furthest point sampling in ablation studies to ensure consistent comparison. We also evaluated the generalizability of our method using real-world data captured by an Intel Realsense D435 RGB-D camera at VGA resolution (640 $\times$ 480 pixels). To mitigate depth camera biases, account for robot dimensions, and align with the deployment setup of \cite{rl_perceptive_locomotion_sr2022}, depth measurements exceeding 1.5m were discarded. 
Consequently, we choose an octree size of 3.2m $\times$ 3.2m $\times$ 3.2m, representing robot-centric environments with a resolution of 0.02m. The input point cloud was generated by back-projecting depth images using camera intrinsic parameters and downsampled to a voxel size of 0.02m. For all experimental groups, we use the following hyperparameter settings: $\theta_{N_\text{out}}$ = 5, $\theta_{r_\text{out}}$ = 5\%, $\theta_{N_\text{in}}$ = 5, $\theta_{e_n}$ = 0.005m$^2$, and $\theta_\text{coplane}$ = 0.85.

\textbf{Evaluation Metrics:} We evaluate the plane directional error $\alpha$ using the mean angular difference between the estimated plane normal $\hat{\normal}$ and the ground truth normal $\normal$.
The ratio of the number of segmented planes to the ground truth $r_p$ and the average ratio of missing points $r_m$ are used to further evaluate the segmentation quality.

\subsection{Ablation Study}

\subsubsection{Number of Sampling Points}
To assess the impact of point numbers on the performance of our method, we conducted experiments using varying point counts: 2048, 4096, and 8192 points sampled from the generated synthetic dataset. As shown in Table~\ref{tbl:as_point_numbers}, the choice of the point number balances the segmentation quality and efficiency. Inadequate sampling, such as using only 2048 points, resulted in detecting only 66\% of the correct planes, with more than 25\% missed points.
Undetected planes primarily occurred for planes of small but feasible size, where insufficient sampling points prevented the satisfaction of Criterion 2 during traversal.
Conversely, excessive points reduced the efficiency of pointwise segmentation, resulting in inference times surpassing 23ms for point quantities surpassing 8192, as detailed in Table~\ref{tbl:inference_time}.
Based on the experimental results, optimal performance was observed with point counts ranging from 4000 to 8000.

\begin{table}[htbp]
\centering
\caption{\textbf{Ablation study on the number of sampling points and effectiveness of the pointwise classification (PC)}. $\downarrow$ means the lower value is better. $\rightarrow 100$ means a value closer to 100 is better.}
\begin{tabular}{l|l|ccc} 
\toprule
\multicolumn{2}{l|}{Point Number} & 2048   & 4096   & 8192      \\ 
\midrule
\midrule
\multirow{3}{*}{with PC}   & $\alpha$ $(^\circ)$ $\downarrow$   & 0.03  & 0.01  & 0.00    \\ 
                           & $r_p$ $\rightarrow 100$        & 66   & 97   & 100      \\
                           & $r_m$ (\%)  $\downarrow$    & 25.2     & 2.08     & 0.00       \\ 
\midrule
\multirow{3}{*}{w/o PC}    & $\alpha$ $(^\circ)$ $\downarrow$   & 9.17 & 8.02 & 12.18  \\
                           & $r_p$ $\rightarrow 100$       & 111 & 122   & 138      \\
                           & $r_m$ (\%)  $\downarrow$     & 6.81     & 0.05     & 0.00   \\ 
\bottomrule
\end{tabular}

\label{tbl:as_point_numbers}
\vspace{-5mm}
\end{table}
\begin{table}[htbp]
\centering
\caption{Inference time for pointwise classification.}
\begin{tabular}{l|cccc}
\toprule
Point Number      & 2048 & 4096 & 8192 \\
\midrule
\midrule
Inference Time (ms)  & 6    & 10   & 23   \\
\bottomrule
\end{tabular}
\label{tbl:inference_time}
\end{table}

\subsubsection{Effectiveness of the Pointwise Classification (PC)} 
\begin{figure}[htbp]
	\centering
	\begin{minipage}{0.13\linewidth}
	    \small with PC
	\end{minipage}
	\begin{minipage}{0.25\linewidth}
		\centering
		\includegraphics[width=\linewidth, trim = 150 0 150 100, clip]{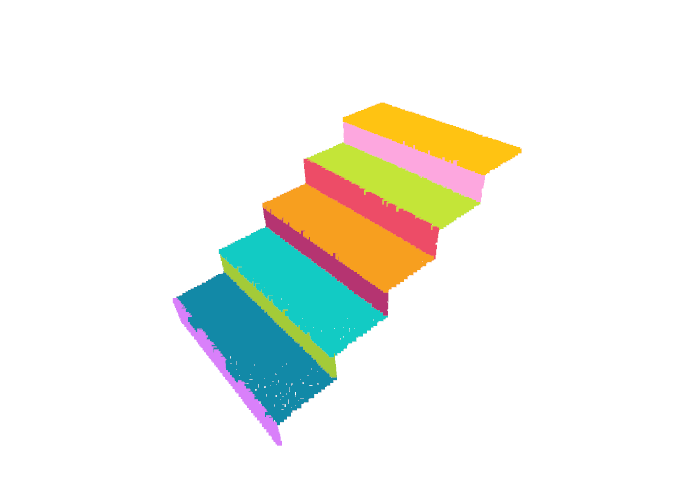}
	\end{minipage}
	\begin{minipage}{0.25\linewidth}
		\centering
		\includegraphics[width=\linewidth, trim = 150 0 150 100, clip]{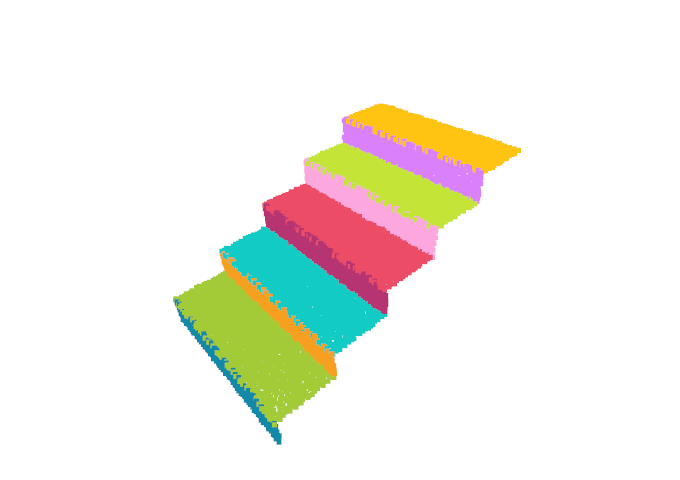}
	\end{minipage}

	
	\begin{minipage}{0.13\linewidth}
	    \small w/o PC
	\end{minipage}
	\begin{minipage}{0.25\linewidth}
		\centering
		\includegraphics[width=\linewidth, trim = 150 0 150 100, clip]{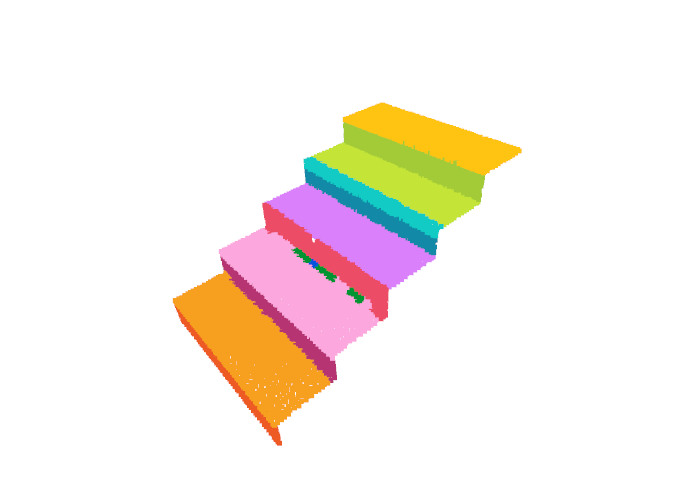}
	\end{minipage}
	\begin{minipage}{0.25\linewidth}
		\centering
		\includegraphics[width=\linewidth, trim = 150 0 150 100, clip]{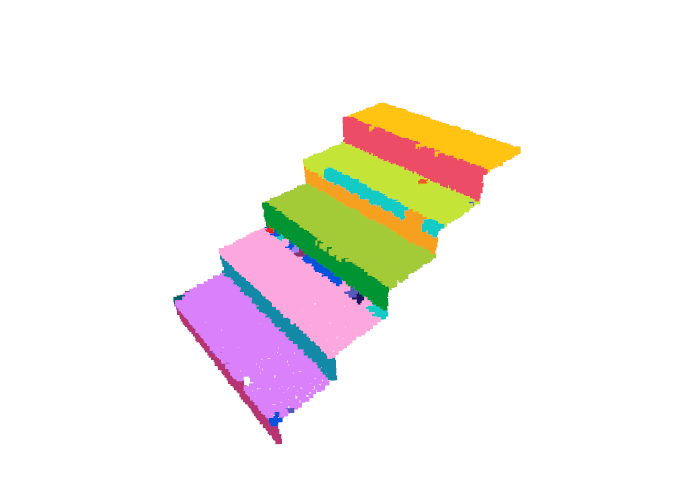}
	\end{minipage}

    \begin{minipage}{0.15\linewidth}
    \centering
    \textbf{}
	\end{minipage}
	\begin{minipage}{0.25\linewidth}
		\centering
		\small noise = 0.0
	\end{minipage}
	\begin{minipage}{0.25\linewidth}
		\centering
		\small noise = 0.3
	\end{minipage}

	\caption{\footnotesize Comparison of the results of plane segmentation with pointwise classification (PC) and without PC at two different noise levels on synthetic data that contains 10 planes. The results using PC both achieve the ground truth, but the results without PC contain 13 and 19 planes, respectively.}
    \label{fig:ablation_stairs_noise}

    \vspace{-3mm}
\end{figure}

To assess the impact of the PC module on improving multi-resolution plane segmentation, we conducted experiments comparing the performance of our method with and without the PC module, using the original synthetic dataset. The dataset was devoid of noise to ensure a fair comparison. In cases where the PC module was absent, we directly applied PCA to each node, as the criteria for node division were inapplicable. As shown in Table~\ref{tbl:as_point_numbers}, using PC significantly reduced the mean plane directional error $\alpha$ to nearly zero and brought the number of segmented planes closer to the ground truth. These results indicate a notable enhancement in the quality of segmentation. We further provide a visualization of the segmentation quality in Figure~\ref{fig:ablation_stairs_noise}, specifically under the condition where the noise level is set to 0.0. This demonstrates our method's enhanced capability in handling plane borders when incorporating the proposed PC module.

\subsubsection{Noise Robustness of the Proposed Method}
\label{sec:as_noise}
To demonstrate the noise robustness of the proposed method, we introduced different levels of pointwise noise $\mathcal{N} \sim (0, \sigma_i^2)$ into the input point cloud, with $\sigma_i$ ranging from $0.0$ to $0.3$. For a fair comparison, we utilized furthest point sampling to sample 8192 points in this study. The corresponding results are presented in Table~\ref{tbl:as_noise_robustness}. We observed that as the noise level increased from $0.0$ to $0.3$, $\alpha$ increased by only $1.36^\circ$, and $r_p$ increased by only $2\%$ when employing PC. In contrast, these metrics exhibited an increase of $2.8^\circ$ and $15\%$, respectively, in the absence of PC. These results also show that PC improves the noise-robust qualities. The corresponding visualization is provided in Figure~\ref{fig:ablation_stairs_noise}.
\begin{table}[htbp]
\centering
\caption{\textbf{Ablation study on noise robustness.} $\downarrow$ means the lower value is better. $\rightarrow 100$ means a value closer to 100 is better.}
 \begin{tabular}{l|l|cccc}
\toprule
\multicolumn{2}{l|}{Noise Level (cm)} & 0.0   & 0.1   & 0.2   & 0.3   \\ 
\midrule
\midrule
\multirow{3}{*}{with PC}   & $\alpha$ $(^\circ)$ $\downarrow$   & 0.00  & 0.10  & 0.26  & 1.36  \\ 
                           & $r_p$ $\rightarrow 100$       & 100   & 100   & 101   & 102   \\
                           & $r_m$ (\%)  $\downarrow$    & 0.00     & 0.00     & 0.00     & 0.08  \\ 
\midrule
\multirow{3}{*}{w/o PC}    & $\alpha$ $(^\circ)$ $\downarrow$   & 12.18 & 12.62 & 13.63 & 14.98 \\
                           & $r_p$ $\rightarrow 100$       & 138 & 144   & 152   & 153   \\
                           & $r_m$ (\%)  $\downarrow$     & 0.00     & 0.00     & 0.00     & 0.04 \\ 
\bottomrule

\end{tabular}
\label{tbl:as_noise_robustness}
\vspace{-2mm}
\end{table}

\subsection{Evaluation on Real Data}
\label{evaluation_real_data}
\begin{figure*}[htbp]
	\centering
        \includegraphics[width=\linewidth, trim = 0 0 0 0, clip]{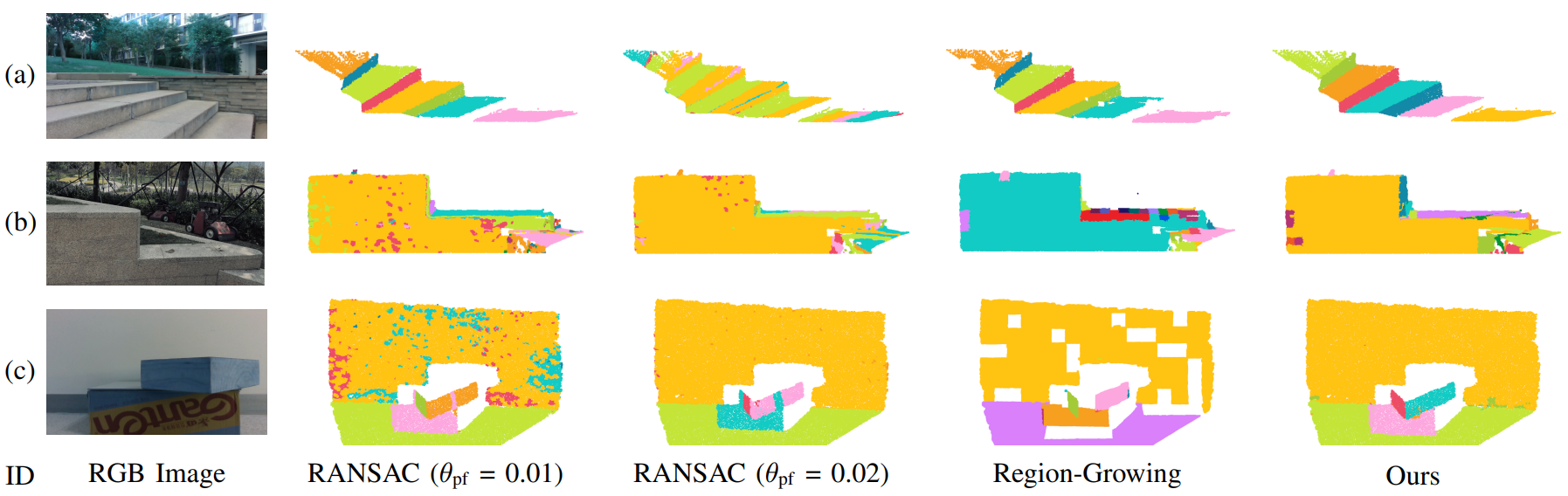}
	\caption{\textbf{Qualitative evaluation results on real data.} We evaluated the proposed method across three diverse terrain scenarios: (a) a stair, (b) grass with a vertical wall, and (c) an indoor composite scene. Depth measurements exceeding 1.5m were disregarded. In each group, the left RGB image displays the ground truth, while the right columns present results for each experimental group, with distinct colors denoting individual planes.}
\label{fig:result_real_data}
\vspace{-5mm}
\end{figure*}
 As shown in Figure~\ref{fig:result_real_data}, we evaluated the proposed method in three scenarios with different terrains: (a) a stair, (b) grass with a vertical wall, and (c) an indoor composite scene. In particular, (c) is specifically designed to evaluate the maximum point numbers per frame, with the camera positioned approximately 1.5 meters from the wall. We recorded several metrics, including the point number of the original point cloud ($N_m$), the point number after down-sampling $N_d$, the time required for down-sampling $t_d$, octree construction time $t_b$, pointwise classification time $t_c$, and traversal time $t_s$. All time-related measurements are reported in milliseconds (ms). The results indicate that our method consistently achieves frame rates exceeding 35FPS across various scenarios.
\begin{table}[htbp]
\vspace{-2mm}
\centering
\caption{Details about the real data.}
\begin{tabular}{c|cc|cccc|c}
\toprule
ID & $N_m$ & $N_d$ & $t_d$ & $t_c$ & $t_b$ & $t_s$ & Speed (FPS) \\ 
\midrule
\midrule
a     & 105033          & 5854        & 3.0                 & 18.8               & 1.4        & 0.6      & 42.0\\ 
b     & 157418          & 6653        & 4.4                 & 20.7               & 2.1        & 0.5      & 36.1 \\ 
c     & 361516          & 4197        & 10.0                & 13.9               & 0.8        & 0.5      & 39.7  \\ 
\bottomrule
\end{tabular}
\label{tbl:real_data}
\vspace{-5mm}
\end{table}

\subsection{Comparison with Existing Methods}
We also compared our method with the existing methods. We not only compared the scores of all the methods under different metrics using the same synthetic data as Section~\ref{sec:as_noise} but also visually compared the segmentation results on real data. A detailed description of each comparison group is summarized as follows:

\textbf{1) RANSAC Approach} ~\cite{ransac_bipedal_iros2021}: We conducted experiments using the RANSAC approach in two groups, where we set the plane fitting thresholds ($\theta_\text{pf}$) to be 0.01m and 0.02m, respectively. Additionally, we set the maximum number of iterations to 1000 and excluded plane candidates with fewer than 50 points.

\textbf{2) Region-Growing Approach} ~\cite{cape_plane_extraction_2018iros, zhixu_plane_extraction_2022iros}: For this approach, we set the threshold for planar region identification to be $0.01\text{m}^2$ and the threshold for plane merging to 0.86. When evaluating on real data, we use depth image patches of size $20 \times 20$ pixels. Notably, we employed 3D region-growing with a voxel size of 0.05m for synthetic data consisting solely of point clouds to conduct the experiments.

\textbf{Results on synthetic data:} The average processing times for the RANSAC, region-growing, and our method are 3.1FPS, 238.7FPS, and 35.3FPS, respectively. The results in Table~\ref{tbl:comparison_synthetic} indicate the robustness of all three methods to noise. However, the region-growing approach leads to more missed points, resulting in losses of resolutions. Compared to existing methods, our approach balances efficiency and accuracy, ensuring more practical applicability.
\begin{table}[htbp]
\centering
\caption{\textbf{Comparison with existing methods on synthetic data.} $\downarrow$ means the lower value is better. $\rightarrow 100$ means a value closer to 100 is better.}
\begin{tabular}{l|l|cccc}
\toprule
\multicolumn{2}{l|}{Noise Level (cm)} & 0.05   & 0.10   & 0.15   & 0.20   \\ 
\midrule
\midrule
\multirow{3}{*}{RANSAC ($\theta_\text{pf}$ = 0.01)}   & $\alpha$ $(^\circ)$ $\downarrow$   & 0.08  & 0.11  & 0.19  & 0.21  \\ 
                           & $r_p$ $\rightarrow 100$       & 100   & 100   & 100   & 100   \\
                           & $r_m$ (\%)  $\downarrow$    & 0.00     & 0.04     & 0.05     & 0.07  \\ 
\midrule
\multirow{3}{*}{Region-Growing}    & $\alpha$ $(^\circ)$ $\downarrow$   & 0.28 & 0.29 & 0.34 & 0.38 \\
                           & $r_p$ $\rightarrow 100$       & 100 & 102   & 102   & 103   \\
                           & $r_m$ (\%)  $\downarrow$     & 0.72     & 0.75     & 0.77    & 0.82 \\ 
\midrule
\multirow{3}{*}{Ours}   & $\alpha$ $(^\circ)$ $\downarrow$   & 0.04  & 0.10  & 0.17  & 0.26  \\ 
                           & $r_p$ $\rightarrow 100$       & 100   & 100   & 101   & 101   \\
                           & $r_m$ (\%)  $\downarrow$    & 0.00     & 0.00     & 0.00     & 0.00  \\ 
\bottomrule
\end{tabular}
\label{tbl:comparison_synthetic}
\vspace{-1mm}
\end{table}

\textbf{Results on real data:} 
 As shown in Figure~\ref{fig:result_real_data}, the RANSAC approach exhibits sensitivity to hyperparameter settings. Setting $\theta_\text{pf}$ to 0.02m yields improved results in scenes (b) and (c) but fails to segment planes in scene (a) accurately. The region-growing approach consistently results in inaccurate plane boundaries and resolution loss. Compared to existing methods, our approach demonstrates robustness across hyperparameter settings, reduces resolution loss, enhances the accuracy of plane boundaries, and ultimately yields superior outcomes.
\section{Conclusion}
This paper introduced a novel multi-resolution method for efficiently extracting planar regions from unordered point clouds. Our approach pre-processes the point cloud with a deep learning approach, then uses an octree-based divide-and-conquer strategy for multi-resolution plane segmentation. We provide an in-depth analysis of point distribution within each octree's node, resulting in accurate and efficient plane segmentation. Extensive experiments demonstrated the effectiveness of our method in segmenting complex terrains in real-world scenarios, showcasing its robustness to noise and real-time performance (over 35FPS). In future research, we aim to use the segmented planes from multiple frames to construct robot-centric maps for perceptive locomotion.

\newpage
\bibliographystyle{ieeetr}
\bibliography{root}


\end{document}